\def\year{2018}\relax
\documentclass[letterpaper]{article} 
\usepackage{aaai18}  
\usepackage{times}  
\usepackage{helvet}  
\usepackage{courier}  
\usepackage{url}  
\usepackage{graphicx}  
\usepackage{amsfonts}
\usepackage{amsmath}
\usepackage{booktabs} 
\usepackage{adjustbox}
\usepackage{caption}
\usepackage{subcaption}
\usepackage{enumitem}
\usepackage[utf8]{inputenc} 
\usepackage[T1]{fontenc}    

\usepackage{tikz}
\usetikzlibrary{arrows, positioning, shapes, fit}
\newcommand{\ignore}[1]{}

\newcommand{\ie}{\textit{i.e.}}
\newcommand{\nop}[1]{}
\newtheorem{definition}{Definition}

\begin{document}
%
\title{Open-World Knowledge Graph Completion}

\author{Baoxu Shi\quad{}Tim Weninger\\
University of Notre Dame\\
\{bshi,tweninge\}@nd.edu\\
}
\maketitle
\begin{abstract}
Knowledge Graphs (KGs) have been applied to many tasks including Web search, link prediction, recommendation, natural language processing, and entity linking. However, most KGs are far from complete and are growing at a rapid pace. To address these problems, Knowledge Graph Completion (KGC) has been proposed to improve KGs by filling in its missing connections. Unlike existing methods which hold a closed-world assumption, \ie, where KGs are fixed and new entities cannot be easily added, in the present work we relax this assumption and propose a new open-world KGC task. As a first attempt to solve this task we introduce an open-world KGC model called ConMask. This model learns embeddings of the entity's name and parts of its text-description to connect unseen entities to the KG. To mitigate the presence of noisy text descriptions, ConMask uses a relationship-dependent content masking to extract relevant snippets and then trains a fully convolutional neural network to fuse the extracted snippets with entities in the KG. Experiments on large data sets, both old and new, show that ConMask performs well in the open-world KGC task and even outperforms existing KGC models on the standard closed-world KGC task. 
\end{abstract}

\section{Introduction}

Knowledge Graphs (KGs) are a special type of information network that represents knowledge using RDF-style triples $\langle h$, $r$, $t\rangle$, where $h$ represents some head entity and $r$ represents some relationship that connects $h$ to some tail entity $t$. In this formalism a statement like ``\textit{Springfield is the capital of Illinois}'' can be represented as $\langle$\textsf{Springfield}, \textsf{capitalOf}, \textsf{Illinois}$\rangle$. Recently, a variety of KGs, such as DBPedia~\cite{Lehmann2014}, and ConceptNet~\cite{Speer2017}, have been curated in the service of fact checking~\cite{Shi2016fc}, question answering~\cite{Lukovnikov2017neural}, entity linking~\cite{Hachey2013}, and for many other tasks~\cite{Nickel2016review}. Despite their usefulness and popularity, KGs are often noisy and incomplete. For example, DBPedia, which is generated from Wikipedia's infoboxes, contains $4.6$ million entities, but half of these entities contain less than 5 relationships. 

Based on this observation, researchers aim to improve the accuracy and reliability of KGs by predicting the existence (or probability) of relationships. This task is often called Knowledge Graph Completion (KGC). Continuing the example from above, suppose the relationship \textsf{capitalOf} is missing between \textsf{Indianapolis} and \textsf{Indiana}; the KGC task might predict this missing relationship based on the topological similarity between this part of the KG and the part containing \textsf{Springfield} and \textsf{Illinois}.

Progress in vector embeddings originating with word2vec has produced major advancements in the KGC task. Typical embedding-based KGC algorithms like TransE~\cite{Bordes2013} and others learn low-dimensional representations (\ie, embeddings) for entities and relationships using topological features. These models are able to predict the existence of missing relationships thereby ``completing'' the KG. 

Existing KGC models implicitly operate under the \emph{Closed-World Assumption}~\cite{reiter1978closed} in which all entities and relationships in the KG cannot be changed -- only discovered. We formally define the Closed-word KGC task as follows:

\begin{definition}\label{def:closed-kgc}
Given an incomplete Knowledge Graph $\mathcal{G}=(\mathbf{E},\mathbf{R},\mathbf{T})$, where $\mathbf{E}$, $\mathbf{R}$, and $\mathbf{T}$ are the entity set, relationship set, and triple set respectively, \textbf{Closed-World Knowledge Graph Completion} completes $\mathcal{G}$ by finding a set of missing triples $\mathbf{T^\prime}=\{\langle h,r,t\rangle | h\in \mathbf{E}, r \in \mathbf{R}, t \in \mathbf{E}, \langle h,r,t\rangle \notin \mathbf{T}\}$ in the incomplete Knowledge Graph $\mathcal{G}$.
\end{definition}

Closed-world KGC models heavily rely on the connectivity of the existing KG and are best able to predict relationships between existing, well-connected entities. Unfortunately, because of their strict reliance on the connectivity of the existing KG, closed-world KGC models are unable to predict the relationships of poorly connected or new entities. Therefore, we assess that closed-world KGC is most suitable for fixed or slowly evolving KGs.

However, most real-world KGs evolve quickly with new entities and relationships being added by the minute. For example, in the 6 months between DBPedia's October 2015 release and its April 2016 release $36,340$ new English entities were added -- a rate of $200$ new entities per day. Recall that DBPedia merely tracks changes to Wikipedia infoboxes, so these updates do not include newly added articles without valid infobox data. Because of the accelerated growth of online information, repeatedly re-training closed-world models every day (or hour) has become impractical.

In the present work we borrow the idea of open-world assumption from probabilistic database literature~\cite{ceylan2016open} and relax the closed-world assumption to develop an \emph{Open-World Knowledge Graph Completion} model capable of predicting relationships involving unseen entities or those entities that have only a few connections. Formally we define the open-world KGC task as follows:

\begin{definition}\label{def:open-kgc}
Given an incomplete Knowledge Graph $\mathcal{G}=(\mathbf{E},\mathbf{R},\mathbf{T})$, where $\mathbf{E}$, $\mathbf{R}$, and $\mathbf{T}$ are the entity set, relationship set, and triple set respectively, \textbf{Open-World Knowledge Graph Completion} completes $\mathcal{G}$ by finding a set of missing triples $\mathbf{T^\prime}=\{\langle h,r,t\rangle | \langle h,r,t\rangle \notin \mathbf{T}, h \in \mathbf{E}^i, t\in \mathbf{E}^i, r \in \mathbf{R} \}$ in the incomplete Knowledge Graph $\mathcal{G}$ where $\mathbf{E}^i$ is an entity superset.
\end{definition}

In Defn.~\ref{def:open-kgc} we relax the constraint on the triple set $\mathbf{T^\prime}$ so that triples in $\mathbf{T^\prime}$ can contain entities that are absent from the original entity set $\mathbf{E}$. 

Closed-world KGC models learn entity and relationship embedding vectors by updating an initially random vector based on the KG's topology. Therefore, any triple $\langle h,r,t\rangle\in\mathbf{T^\prime}$ such that $h\notin \mathbf{E}$ or $t\notin \mathbf{E}$ will only ever be represented by its initial random vector because its absence does not permit updates from any inference function. In order to predict the missing connections for unseen entities, it is necessary to develop alternative features to replace the topological features used by closed-world models.

Text content is a natural substitute for the missing topological features of disconnected or newly added entities. Indeed, most KGs such as FreeBase~\cite{Bollacker2008}, DBPedia~\cite{Lehmann2014}, and SemMedDB~\cite{Kilicoglu2012} were either directly extracted from~\cite{Lin2016RE,ji2011knowledge}, or are built in parallel to some underlying textual descriptions. However, open-world KGC differs from the standard information extraction task because 1) Rather than extracting triples from a large text corpus, the goal of open-world KGC is to discover missing relationships; and 2) Rather than a pipeline of independent subtasks like Entity Linking~\cite{Francis2016} and Slotfilling~\cite{Liu2016}, etc., open-world KGC is a holistic task that operates as a single model. 

Although it may seem intuitive to simply include an entity's description into an existing KGC model, we find that learning useful vector embeddings from unstructured text is much more challenging than learning topology-embeddings as in the closed-world task. First, in closed-world KGC models, each entity will have a unique embedding, which is learned from its directly connected neighbors; whereas open-world KGC models must fuse entity embeddings with the word embeddings of the entity's description. These word embeddings must be updated by entities sharing the same words regardless of their connectivity status. Second, because of the inclusion of unstructured content, open-world models are likely to include noisy or redundant information.

With respect to these challenges, the present work makes the following contributions:

\begin{enumerate}
    \item We describe an open-world KGC model called \textbf{ConMask} that uses relationship-dependent content masking to reduce noise in the given entity description and uses fully convolutional neural networks (FCN) to fuse related text into a relationship-dependent entity embedding.
    \item We release two new Knowledge Graph Completion data sets constructed from DBPedia and Wikipedia for use in closed-world and open-world KGC evaluation.
\end{enumerate}

Before introduce the ConMask model, we first present preliminary material by describing relevant KGC models. Then we describe the methodology, data sets, and a robust case study of closed-world and open-world KGC tasks. Finally, we draw conclusions and offer suggestions for future work.

\section{Closed-World Knowledge Graph Completion}\label{sec:kgc_related_work}

\ignore{
\begin{table}[]
    \caption{Translation function of various closed-world KGC models. $\mathbf{h}$, $\mathbf{r}$, $\mathbf{t}$ are the topological embeddings of the head entity, relationship, and tail entity respectively. $\mathbf{w}_r$ and $\mathbf{M_r}$ are relationship-dependent transformation matrices. $\mathbf{D}_e$ and $\mathbf{D}_r$ are two diagonal transformation matrices and $b_c$ contains learned combination weights.}
    \label{tab:kgc_translation}
    \centering
    \begin{adjustbox}{max width=\linewidth}
    \begin{tabular}{lcccc}
        Model & $f_{c}$ & $f_{h}$ & $f_{r}$ & $f_{t}$ \\
        \midrule
        TransE~\cite{Bordes2013} & $\cdot+\cdot$ & $\mathbf{h}$ & $\mathbf{r}$ & $\mathbf{t}$ \\ 
        TransH~\cite{Wang2014} & $\cdot+\cdot$ & $\mathbf{h} - \mathbf{w}_r^T\mathbf{h}\mathbf{w}_r$ & $\mathbf{r}$ & $\mathbf{t} - \mathbf{w}_r^T\mathbf{t}\mathbf{w}_r$ \\
        TransR~\cite{Lin2015} & $\cdot+\cdot$ & $\mathbf{h}\mathbf{M}_{r}$ & $\mathbf{r}$ & $\mathbf{t}\mathbf{M}_{r}$ \\
        ProjE~\cite{Shi2016proje}  & $\cdot\times\cdot + b_c$ & $\mathbf{D}_{e}\mathbf{h}$ & $\mathbf{D}_{r}\mathbf{r}$ & $\mathbf{t}$ \\
        \bottomrule
    \end{tabular}
    \end{adjustbox}
\end{table}
}

A variety of models have been developed to solve the closed-world KGC task. The most fundamental and widely used model is a translation-based Representation Learning (RL) model called TransE~\cite{Bordes2013}. TransE assumes there exists a simple function that can translate the embedding of the head entity to the embedding of some tail entity via some relationship:

\begin{equation}\label{eq:translation_simple}
    \mathbf{h} + \mathbf{r} = \mathbf{t},
\end{equation}

\ignore{\noindent{}where $\mathbf{h}$, $\mathbf{r}$ and $\mathbf{t}$ are low-dimensional embeddings of head entity, relationship, and tail entity respectively. The definition in Eq.~\ref{eq:translation_simple} addresses one-to-one relationships but ignores the one-to-many and many-to-one relationships that are often found in KGs. For example, in DBPedia the tail entity of a partial triple $\langle$\textsf{Chicago}, \textsf{locatedIn}, \textsf{?}$\rangle$ can be either \textsf{Illinois}, \textsf{United States}, or both. Despite their clear semantic difference, according to Eq.~\ref{eq:translation_simple}, TransE would be trained to represent both \textsf{Illinois} and \textsf{United States} with a similar embedding.}

\noindent{}where $\mathbf{h}$, $\mathbf{r}$ and $\mathbf{t}$ are embeddings of head entity, relationship, and tail entity respectively. Based on this function, many other KGC models improve the expressive power of Eq.~\ref{eq:translation_simple} by introducing more relationship-dependent parameters. TransR~\cite{Lin2015}, for example, augments Eq.~\ref{eq:translation_simple} to $\mathbf{h}\mathbf{M}_{r} + \mathbf{r} = \mathbf{t}\mathbf{M}_{r}$ where $\mathbf{M}_{r}$ is a relationship-dependent entity embedding transformation.

In order to train the KGC models, TransE defines an energy-based loss function as

\begin{equation}\label{eq:mainloss}
\mathcal{L}(\mathbf{T}) = \Sigma_{\langle h,r,t\rangle\in\mathbf{T}}[\gamma + E(\langle h,r,t\rangle) - E(\langle h^\prime, r^\prime, t^\prime\rangle )]_{+},
\end{equation}

\noindent{}where the energy function $E(\langle h,r,t\rangle) = \parallel \mathbf{h} + \mathbf{r} - \mathbf{t}\parallel_{L_{n}}$ measures the closeness of the given triple, $\langle h,r,t\rangle$ is some triple that exists in the triple set $\mathbf{T}$ of an incomplete KG $\mathcal{G}$, and $\langle h^\prime, r^\prime, t^\prime\rangle$ is a ``corrupted'' triple derived by randomly replacing one part of $\langle h,r,t\rangle$ so that it does not exist in $\mathbf{T}$.

In other recent work, ProjE~\cite{Shi2016proje} considered closed-world KGC to be a type of ranking task and applied a list-wise ranking loss instead of Eq.~\ref{eq:mainloss}. Other closed-world models such as PTransE~\cite{Lin2015ptranse} and dORC~\cite{Zhang2017} maintain a simple translation function and use complex topological features like extended-length paths and ``one-relation-circle'' structures to improve predictive performance.

Unlike topology-based models, which have been studied extensively, there has been little work that utilizes text information for KGC. Neural Tensor Networks (NTN)~\cite{Socher2013} uses the averaged word embedding of an entity to initialize the entity representations. DKRL~\cite{Xie2016} uses the combined distance between topology-embeddings and text-embeddings as its energy function. Jointly~\cite{Xu2016} combines the topology-embeddings and text-embeddings first using a weighted sum and then calculates the $L_{n}$ distance between the translated head entity and tail entity. However, gains in predictive performance from these joint-learning models are rather small compared to advances in topology-based models.

Furthermore, the aforementioned models are all closed-world KGC models, which can only learn meaningful representations for entities that are present during training and are well connected within the KG. These models have no mechanism by which new entities can be connected with the existing KG as required in open-world KGC.

In the present work, we present an open-world KGC model called ConMask that uses primarily text features to learn entity and relationship embeddings. Compared to topology-based and joint-learning models, ConMask can generate representations for unseen entities if they share the same vocabulary with entities seen during training. To properly handle one-to-many and many-to-one relationships, we also apply a relationship-dependent content masking layer to generate entity embeddings. 

\section{ConMask: A Content Masking Model for Open-World KGC}\label{sec:kgc_conmask}

In this section we describe the architecture and the modelling decisions of the ConMask model. To illustrate how this model works, we begin by presenting an actual example as well as the top-ranked target entity inferred by the ConMask model: 

\noindent{}\textbf{Example Task}: Complete triple $\langle$\textsf{Ameen Sayani}, \textsf{residence}, \textsf{?} $\rangle$, where \textsf{Ameen Sayani} is absent from the KG.

\noindent{}\textbf{Snippet of Entity Description}: ``\textit{... \textbf{Ameen Sayani} was introduced to All India Radio, \textbf{Bombay}, by his brother Hamid Sayani. Ameen participated in English programmes there for ten years ...}''.

\noindent{}\textbf{Predicted Target Entity}: \textsf{Mumbai}.

In this example, if a human reader were asked to find the residence of Ameen Sayani, a popular radio personality in India, from the entity description, then the human reader is unlikely to read the entire text from beginning to end. Instead, the reader might skim the description looking for contextual clues such as family or work-related information. Here, Ameen's workplace All India Radio is located in Bombay, so the human reader may infer that Ameen is a resident of Bombay. A human reader may further reason that because Bombay has recently changed its name to Mumbai, then \textsf{Mumbai} would be the (correct) target entity.

Here and throughout the present work, we denote the missing entity as the \emph{target} entity, which can be either the head or the tail of a triple. 

We decompose the reasoning process described above into three steps: 1) Locating information relevant to the task, 2) Implicit reasoning based on the context and the relevant text, and 3) Resolving the relevant text to the proper target entity. The ConMask model is designed to mimic this process. Thus, ConMask consists of three components: 
\begin{enumerate}
    \item Relationship-dependent content masking, which highlights words that are relevant to the task,
    \item Target fusion, which extracts a target entity embedding from the relevant text, and
    \item Target entity resolution, which chooses a target entity by computing a similarity score between target entity candidates in the KG, the extracted entity embeddings, and other textual features.
\end{enumerate}

\begin{figure}[t]
    \centering
    \begin{tikzpicture}

\tikzstyle{inputbox} = [rectangle, row sep=50, column sep=2, text centered, draw=black, fill=white]
\tikzstyle{word} = [circle, draw=black, fill=gray!50, minimum width=8, minimum height=8]
\tikzstyle{skip} = [rectangle, inner sep=1]
\tikzstyle{layer} = [rectangle, minimum width=1, minimum height=2, draw=black, align=center]
\tikzstyle{arrow} = [thick, ->, >=stealth]
\tikzstyle{feature} = [inner sep=1, circle, draw=black, minimum width=10, minimum height=10]

\tikzstyle{lookup} = [rectangle, minimum width=10, minimum height=10, draw=black, fill=red!20]
\tikzstyle{avg} = [inner sep=1, star, star points=4, minimum width=12, minimum height=12, draw=black, fill=green!20]
\tikzstyle{mask} = [ inner sep=2, regular polygon, regular polygon sides=5, minimum width=12, minimum height=12, draw=black, fill=yellow!20]
\tikzstyle{fusion} = [inner sep=0, star, star points=5, minimum width=12, minimum height=12, draw=black, fill=cyan!20]


\matrix[name=head_name, inputbox, label={left:$\psi(h)$}, label={\small Head Entity Name},] {
	\node[word]{}; &
	\node[skip]{\ldots}; &
	\node[word]{}; \\
};

\matrix[name=head_content, inputbox, label={left:$\phi(h)$}, label={\small Head Entity Description}, below= of head_name] {
	\node[word]{}; &
	\node[word]{}; &
	\node[skip]{\ldots}; &
	\node[word]{}; &
    \node[word]{}; \\
};

\matrix[name=rel_name, inputbox, label={left:$\psi(r)$}, label={\small Relationship Name}, below= of head_content] {
	\node[word]{}; &
	\node[skip]{\ldots}; &
	\node[word]{}; \\
};

\matrix[name=tail_content, inputbox, label={left:$\phi(t)$}, label={\small Tail Entity Description}, below= of rel_name] {
	\node[word]{}; &
	\node[word]{}; &
	\node[skip]{\ldots}; &
	\node[word]{}; &
    \node[word]{}; \\
};

\matrix[name=tail_name, inputbox, label={left:$\psi(t)$}, label={\small Tail Entity Name}, below= of tail_content] {
	\node[word]{}; &
	\node[skip]{\ldots}; &
	\node[word]{}; \\
};

\node[name=head_content_lookup, lookup, right= .3 of head_content] {};
\node[name=head_name_lookup, lookup] at (head_name -| head_content_lookup) {};
\node[name=rel_name_lookup, lookup] at (rel_name -| head_content_lookup) {};
\node[name=tail_content_lookup, lookup] at (tail_content -| rel_name_lookup) {};
\node[name=tail_name_lookup, lookup] at (tail_name -| tail_content_lookup)  {};

\draw[arrow] (head_name)--(head_name_lookup);
\draw[arrow] (head_content)--(head_content_lookup);
\draw[arrow] (rel_name)--(rel_name_lookup);
\draw[arrow] (tail_content)--(tail_content_lookup);
\draw[arrow] (tail_name)--(tail_name_lookup);

\node[name=head_mask, mask, right= 0.4 of head_content_lookup] {$\tau$};
\node[name=tail_mask, mask, right = 0.4 of tail_content_lookup] {$\tau$};

\draw[arrow] (head_content_lookup)--(head_mask);
\draw[arrow] (tail_content_lookup)--(tail_mask);

\node[name=head_name_avg, avg, right = 0.4 of head_name_lookup] {$\eta$};

\node[name=rel_avg, avg, right = 0.4 of rel_name_lookup] {$\eta$};
\node[name=head_content_avg, avg, below right = .7 and .5 of head_mask] {$\eta$};
\node[name=tail_content_avg, avg, above right = .7 and .5 of tail_mask] {$\eta$};
\node[name=tail_name_avg, avg] at(tail_name_lookup -| tail_mask) {$\eta$};

\draw[arrow] (head_name_lookup)--(head_name_avg);
\draw[arrow] (head_content_lookup)--(head_content_avg);
\draw[arrow] (tail_content_lookup)--(tail_content_avg);
\draw[arrow] (rel_name_lookup)--(rel_avg);
\draw[arrow] (tail_name_lookup)--(tail_name_avg);

\node[name=head_fusion, fusion, right = 1.0 of head_mask] {$\xi$};
\node[name=tail_fusion, fusion, right = 1.0 of tail_mask] {$\xi$};

\draw[arrow] (head_mask)--(head_fusion);
\draw[arrow] (rel_name_lookup)--(head_fusion);
\draw[arrow] (tail_mask)--(tail_fusion);
\draw[arrow] (rel_name_lookup)--(tail_fusion);


\node[name=f4, feature, fill=magenta!50, right = 2.0 of rel_avg] {$\theta_4$};

\node[name=f3, feature, fill=magenta!50, above = .5 of f4] {$\theta_3$};
\node[name=f2, feature, fill=magenta!50, above = .5 of f3] {$\theta_2$};
\node[name=tail_content_head_name, feature, fill=cyan!50, above= .5 of f2] {$\theta_1$};

\node[name=f5, feature, fill=magenta!50, below = .5 of f4] {$\theta_5$};
\node[name=f6, feature, fill=magenta!50, below = .5 of f5] {$\theta_6$};

\node[name=head_content_tail_name, feature, fill=cyan!50, below = .5 of f6] {$\theta_7$};

\draw[arrow] (head_fusion)--(head_content_tail_name);
\draw[arrow] (tail_name_avg)--(head_content_tail_name);
\draw[arrow] (tail_fusion)--(tail_content_head_name);
\draw[arrow] (head_name_avg)--(tail_content_head_name);

\draw[arrow] (head_content_avg)--(f4);
\draw[arrow] (tail_content_avg)--(f4);

\draw[arrow] (head_fusion)--(f2);
\draw[arrow] (tail_fusion)--(f2);

\draw[arrow] (rel_avg)--(f3);
\draw[arrow] (head_name_avg)--(f3);

\draw[arrow] (rel_avg)--(f5);
\draw[arrow] (tail_name_avg)--(f5);

\draw[arrow] (head_name_avg)--(f6);
\draw[arrow] (tail_name_avg)--(f6);



\node[name=output, feature, right = 1.2 of f4] {};

\draw[arrow] (tail_content_head_name)--(output) node [midway, fill=white] {\small$\mathbf{w}_1$};
\draw[arrow] (f2)--(output) node [midway, fill=white] {\small$\mathbf{w}_2$};
\draw[arrow] (f3)--(output) node [midway, fill=white] {\small$\mathbf{w}_3$};
\draw[arrow] (f4)--(output) node [midway, fill=white] {\small$\mathbf{w}_4$};
\draw[arrow] (f5)--(output) node [midway, fill=white] {\small$\mathbf{w}_5$};
\draw[arrow] (f6)--(output) node [midway, fill=white] {\small$\mathbf{w}_6$};
\draw[arrow] (head_content_tail_name)--(output) node [midway, fill=white] {\small$\mathbf{w}_7$};




\end{tikzpicture}
    \tikzstyle{inputbox} = [rectangle, row sep=50, column sep=2, text centered, draw=black, fill=white]
\tikzstyle{word} = [circle, draw=black, fill=gray!50, minimum width=10, minimum height=10]
\tikzstyle{skip} = [rectangle]
\tikzstyle{layer} = [rectangle, minimum width=1, minimum height=2, draw=black, align=center]
\tikzstyle{arrow} = [thick, ->, >=stealth]
\tikzstyle{feature} = [inner sep=1, circle, draw=black]
\tikzstyle{lookup} = [rectangle, minimum width=10, minimum height=10, draw=black, fill=red!20]
\tikzstyle{avg} = [inner sep=1, star, star points=4, minimum width=12, minimum height=12, draw=black, fill=green!20]
\tikzstyle{mask} = [ inner sep=2, regular polygon, regular polygon sides=5, minimum width=12, minimum height=12, draw=black, fill=yellow!20]
\tikzstyle{fusion} = [inner sep=0, star, star points=5, minimum width=12, minimum height=12, draw=black, fill=cyan!20]
\tikzstyle{extractive} = [inner sep=1, circle, draw=black, fill=magenta!50, minimum width=12, minimum height=12]
\tikzstyle{contextual} = [inner sep=1, circle, draw=black, fill=cyan!50, minimum width=12, minimum height=12]
\begin{tikzpicture}
\matrix[name=legend, inputbox, row sep=5, column sep=5, draw=white]{
  \node[name=w, word] {}; &
  \node[name=l, lookup] {}; &
  \node[name=a, avg] {$\eta$}; &
  \node[name=m, mask] {$\tau$}; &
  \node[name=f, fusion] {$\xi$}; \\
  \node[align=center, anchor=south] {\small Word}; & 
  \node[align=center] {\small Embedding\\ \small Lookup}; &
  \node[align=center] {\small Semantic\\ \small Averaging}; &
  \node[align=center] {\small Content\\ \small Masking}; & 
  \node[align=center] {\small Target\\ \small Fusion}; &\\
};
\end{tikzpicture}
\begin{tikzpicture}
\matrix[name=legend, inputbox, row sep=5, column sep=5, draw=white]{
  \node[name=e, extractive] {$\theta$}; & 
  \node[name=c, contextual] {$\theta$}; & \\    
  \node[align=center] {\small Contextual\\ \small Feature}; &
  \node[align=center] {\small Extraction\\ \small Feature};\\
};
\end{tikzpicture}
    \caption{Illustration of the ConMask model for Open-World Knowledge Graph Completion.\ignore{ Features $\theta_1$--$\theta_7$ correspond to the similarity functions found in Tab.~\ref{tab:features}.}}
    \label{fig:conmask}
\end{figure}
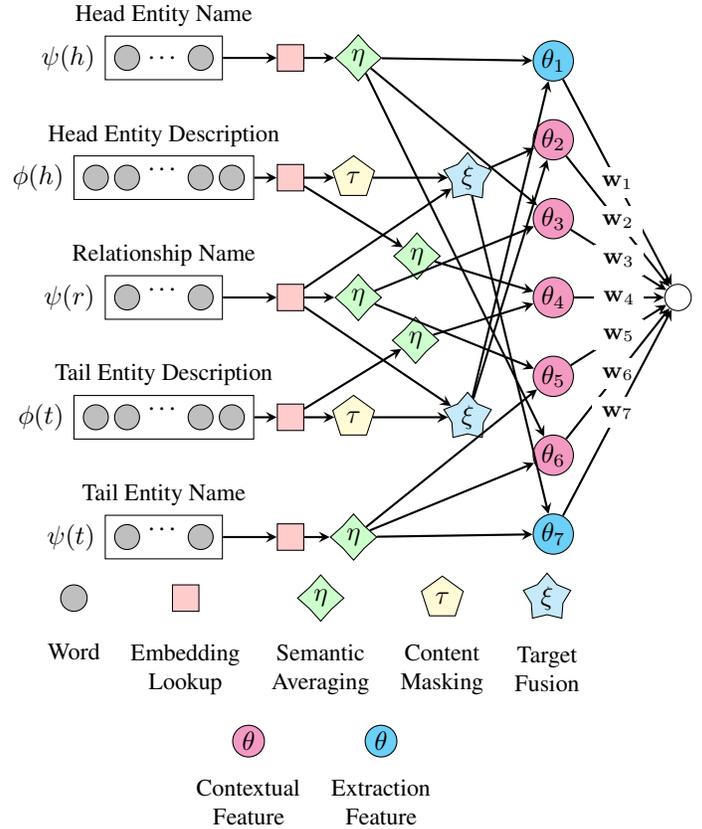

ConMask selects words that are related to the given relationship to mitigate the inclusion of irrelevant and noisy words. From the relevant text, ConMask then uses fully convolutional network (FCN) to extract word-based embeddings. Finally, it compares the extracted embeddings to existing entities in the KG to resolve a ranked list of target entities. The overall structure of ConMask is illustrated in Fig.~\ref{fig:conmask}. Later subsections describe the model in detail.

\subsection{Relationship-Dependent Content Masking}

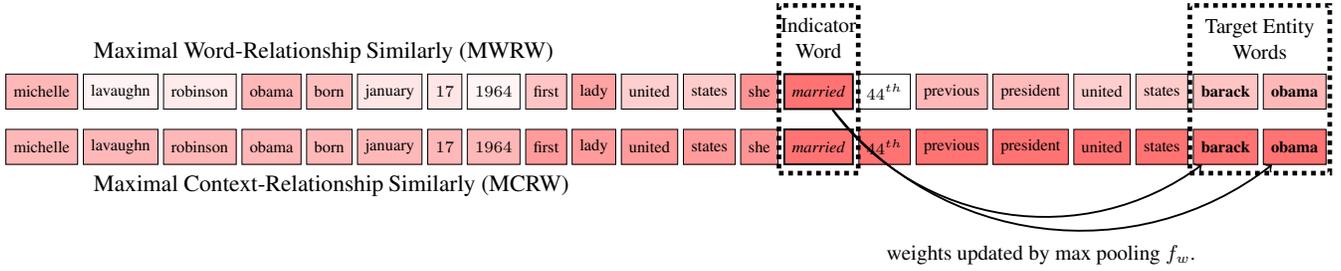
\begin{figure*}[ht]
    \centering
    \begin{adjustbox}{max width=\linewidth}
    \begin{tikzpicture}[every text node part/.style={align=center}, node distance=0.3cm]
\tikzstyle{m} = [rectangle, row sep=8, column sep=2, text centered, draw=white]
\tikzstyle{word} = [rectangle, draw=black, minimum height=15]

\matrix[name=head_name, m] {
\node[word, name=b1, fill=red!28]{\scriptsize michelle}; &
\node[word, name=b2, fill=red!5]{\scriptsize lavaughn}; &
\node[word, name=b3, fill=red!8]{\scriptsize robinson}; &
\node[word, name=b4, fill=red!28]{\scriptsize obama}; &
\node[word, name=b5, fill=red!25]{\scriptsize born}; &
\node[word, name=b6, fill=red!10]{\scriptsize january}; &
\node[word, name=b7, fill=red!10]{\scriptsize $17$}; &
\node[word, name=b8, fill=red!4]{\scriptsize $1964$}; &
\node[word, name=b12, fill=red!23]{\scriptsize first}; &
\node[word, name=b13, fill=red!36]{\scriptsize lady}; &
\node[word, name=b14, fill=red!19]{\scriptsize united}; &
\node[word, name=b15, fill=red!20]{\scriptsize states}; &
\node[word, name=b16, fill=red!42]{\scriptsize she}; &
\node[word, name=b17, thick, fill=red!55]{\scriptsize \textit{married}}; &
\node[word, name=b18, fill=red!0]{\scriptsize $44^{th}$}; &
\node[word, name=b19, fill=red!28]{\scriptsize previous}; &
\node[word, name=b20, fill=red!27]{\scriptsize president}; &
\node[word, name=b21, fill=red!19]{\scriptsize united}; &
\node[word, name=b22, fill=red!20]{\scriptsize states}; &
\node[word, name=b23, fill=red!23]{\scriptsize \textbf{barack}}; &
\node[word, name=b24, fill=red!28]{\scriptsize \textbf{obama}};\\
\node[word, name=a1, fill=red!28]{\scriptsize michelle}; &
\node[word, name=a2, fill=red!28]{\scriptsize lavaughn}; &
\node[word, name=a3, fill=red!28]{\scriptsize robinson}; &
\node[word, name=a4, fill=red!28]{\scriptsize obama}; &
\node[word, name=a5, fill=red!28]{\scriptsize born}; &
\node[word, name=a6, fill=red!28]{\scriptsize january}; &
\node[word, name=a7, fill=red!28]{\scriptsize $17$}; &
\node[word, name=a8, fill=red!28]{\scriptsize $1964$}; &
\node[word, name=a12, fill=red!39]{\scriptsize first}; &
\node[word, name=a13, fill=red!39]{\scriptsize lady}; &
\node[word, name=a14, fill=red!39]{\scriptsize united}; &
\node[word, name=a15, fill=red!39]{\scriptsize states}; &
\node[word, name=a16, fill=red!42]{\scriptsize she}; &
\node[word, name=a17, thick, fill=red!55]{\scriptsize \textit{married}}; &
\node[word, name=a18, fill=red!55]{\scriptsize $44^{th}$}; &
\node[word, name=a19, fill=red!55]{\scriptsize previous}; &
\node[word, name=a20, fill=red!55]{\scriptsize president}; &
\node[word, name=a21, fill=red!55]{\scriptsize united}; &
\node[word, name=a22, fill=red!55]{\scriptsize states}; &
\node[word, name=a23, fill=red!55]{\scriptsize \textbf{barack}}; &
\node[word, name=a24, fill=red!55]{\scriptsize \textbf{obama}};\\
};

\path[->] (b17) edge[thick, bend right=45] (a23);
\path[->] (b17) edge[thick, bend right=45] node [midway, label=below:{\small{weights updated by max pooling $f_w$.}}] {} (a24);

\node[name=wr, right] at (-8.75,1.0) {Maximal Word-Relationship Similarly (MWRW)};
\node[name=cr, right] at (-8.75,-1.0) {Maximal Context-Relationship Similarly (MCRW)};

\node[name=indicator, above= 0.1 of b17, align=center] {\small Indicator\\ \small Word};
\node[dotted, thick, line width=2, draw=black, minimum width=34, minimum height=72] at (2.30,0.45) {};

\node[name=target, above left=.1 and -0.9 of b24] {\small Target Entity\\ \small Words};
\node[dotted, thick, line width=2, draw=black,minimum width=60, minimum height=72] at (8.95,0.45) {};

\end{tikzpicture}
    \end{adjustbox}
    \caption{Relationship-dependent Content Masking heat map for the description of \textsf{Michelle Obama} given relationship type \textsf{spouse}. Stop-words are removed. Higher weights show in darker color.}
    \label{fig:heatmap}
\end{figure*}

In open-world KGC, we cannot rely solely on the topology of the KG to guide our model. Instead, it is natural to consider extracting useful information from text in order to infer new relationships in the KG. The task of extracting relationships among entities from text is often called relation extraction~\cite{mintz2009distant}. Recent work in this area tends to employ neural networks such as CNN~\cite{Xu2016} or abstract meaning representations (AMRs)~\cite{huang2017liberal} to learn a unified kernel to remove noise and extract the relationship-agnostic entity representations. For open-world KGC, it may be possible to create a model with relationship-dependent CNN kernels. But this type of model would significantly increase the number of parameters and may overfit on rare relationships.

In the proposed ConMask model we developed an alternative approach called \textit{relationship-dependent content masking}. The goal is to pre-process the input text in order to select small relevant snippets based on the given relationship -- thereby masking irrelevant text. The idea of content masking is inspired by the attention mechanism used by recurrent neural network (RNN) models~\cite{Chorowski2015}, which is widely applied to NLP tasks. In a typical attention-based RNN model, each output stage of a recurrent cell is assigned an attention score.

In ConMask, we use a similar idea to select the most related words given some relationship and mask irrelevant words by assigning a relationship-dependent similarity score to words in the given entity description. We formally define relationship-dependent content masking as:

\begin{equation}\label{eq:content_masking}
\tau(\phi(e), \psi(r)) = \mathbf{W}_{\phi(e)} \circ f_{w}(\mathbf{W}_{\phi(e)}, \mathbf{W}_{\psi(r)}),
\end{equation}

\noindent{}where $e$ is an entity, $r$ is some relationship, $\phi$ and $\psi$ are the description and name mapping functions respectively that return a word vector representing the description or the name of an entity or relationship. $\mathbf{W}_{\phi(e)} \in \mathbb{R}^{|\phi(e)|\times k}$ is the\ignore{ L$_1$ normalized} description matrix of $e$ in which each row represents a $k$ dimensional embedding for a word in $\phi(e)$ in order, $\mathbf{W}_{\psi(r)} \in \mathbb{R}^{|\psi(r)|\times k}$ is the\ignore{ L$_1$ normalized} name matrix of $r$ in which each row represents a $k$ dimensional embedding for a word in the title of relationship $\psi(r)$, $\circ$ is row-wise product, and $f_{w}$ calculates the masking weight for each row, \ie, the embedding of each word, in $\mathbf{W}_{\phi(e)}$.

The simplest way to generate these weights is by calculating a similarity score between each word in entity description $\phi(e)$ and the words in relationship name $\psi(r)$. We call this simple function Maximal Word-Relationship Weights (MWRW) and define it as:

\begin{equation}\label{eq:mwrw}
    \begin{adjustbox}{max width=0.92\linewidth}
    $f_{w}^{\textrm{MWRW}}\left(\mathbf{W}_{\phi(e)}, \mathbf{W}_{\psi(r)}\right)_{[i]} = \mathsf{max}_j\left(\frac{\sum\limits_m^k \mathbf{W}_{\phi(e)[i,m]} \mathbf{W}_{\psi(r)[j,m]}}{\sqrt{\sum\limits_m^k \mathbf{W}_{\phi(e)[i,m]}^2}\sqrt{\sum\limits_m^k \mathbf{W}_{\psi(r)[j,m]}^2}}\right)$,
    \end{adjustbox}
\end{equation}

\noindent{}where the weight of the $i^{\textrm{th}}$ word in $\phi(e)$ is the largest cosine similarity score between the $i^{\textrm{th}}$ word embedding in $\mathbf{W}_{\phi(e)}$ and the word embedding matrix of $\psi(r)$ in $\mathbf{W}_{\psi(r)}$. 

This function assigns a lower weight to words that are not relevant to the given relationship and assigns higher scores to the words that appear in the relationship or are semantically similar to the relationship. For example, when inferring the target of the partial triple $\langle$\textsf{Michelle Obama}, \textsf{AlmaMater}, \textsf{?}$\rangle$, MWRW will assign high weights to words like \textit{Princeton}, \textit{Harvard}, and \textit{University}, which include the words that describe the target of the relationship. However the words that have the highest scores do not always represent the actual target but, instead, often represent words that are similar to the relationship name itself. A counter-example is shown in Fig.~\ref{fig:heatmap}, where, given the relationship \textsf{spouse}, the word with the highest MWRW score is \textit{married}. Although \textsf{spouse} is semantically similar to \textit{married}, it does not answer the question posed by the partial triple. Instead, we call words with high MWRW weights \emph{indicator words} because the correct target-words are usually located nearby. In the example-case, we can see that the correct target \textit{Barack Obama} appears after the indicator word \textit{married}. 

In order to assign the correct weights to the target words, we improve the content masking by using Maximal Context-Relationship Weights (MCRW) to adjust the weights of each word based on its context:

\begin{equation}\label{eq:mcrw}
    \begin{adjustbox}{max width=0.92\linewidth}
        $f_{w}\left(\mathbf{W}_{\phi(e)}, \mathbf{W}_{\psi(r)}\right)_{[i]} = \max\left(f_{w}^{\textrm{MWRW}}\left(\mathbf{W}_{\phi(e)}, \mathbf{W}_{\psi(r)}\right)_{[i-k_m:i]}\right),$
    \end{adjustbox}
\end{equation}

\noindent{}in which the weight of the $i^{th}$ word in $\phi(e)$ equals the maximum MWRW score of the $i^{th}$ word itself and previous $k_m$ words. From a neural network perspective, the re-weighting function $f_w$ can also be viewed as applying a row-wise max reduction followed by a 1-D max-pooling with a window size of $k_m$ on the matrix product of $\mathbf{W}_{\phi(e)}$ and $\mathbf{W}_{\psi(r)}^{T}$. 

\begin{figure*}[ht]
    \centering
\begin{tikzpicture}[every label/.style={align=center}]
    	\tikzstyle{content} = [rectangle, draw=black, minimum height=50, fill=blue!20, text width=0em]
    	\tikzstyle{nm} = [rectangle, draw=black, minimum height=30, fill=blue!20, text width=0em]
    	\tikzstyle{mat} = [rectangle, draw=black, minimum width=20, minimum height=50, fill=blue!20]
        \tikzstyle{myarrows}=[line width=0.3mm,draw=blue,-triangle 45,postaction={draw, line width=1mm, shorten >=2mm, -}]
        \tikzstyle{arrow} = [thick, ->, >=stealth]
        \tikzstyle{op} = [circle, draw=black, minimum width=5, minimum height=5]

	\node[font=\footnotesize, align=left, text width=1.25cm] at (-0.3,1.3) {Entity Desc.};
    	\node[name=cv, content, minimum width=1, fill=gray!50, label={[xshift=3,yshift=-26]left:\small $k_c$}] {};
    	\node[font=\footnotesize, align=left, text width=1.25cm] at (-0.3,-1.5) {Rel. Name};
    	\node[name=nv, nm, minimum width=1, fill=gray!50, below = of cv, label={[xshift=5,yshift=-16]left:\small$k_n$}] {};
    	\node[name=cembed, content, minimum width=10, right= 0.7 of cv, label=above:\small $200$, label={[xshift=5,yshift=-26]left:\small$k_c$}] {};
    	\draw[myarrows, draw=red!70] (cv)--(cembed)node[near start, above]{\small $l$};
    	\node[name=nembed, nm, minimum width=10, right=0.7 of nv, label=above:\small $200$, label={[xshift=5,yshift=-16]left:\small$k_n$}] {};
    	\draw[myarrows, draw=red!70] (nv)--(nembed) node[near start, above]{\small $l$};
    	
    	\node[name=matmul, op, below right = .2 and 0.3 of cembed] {\small$\times$};
    	\node[name=m, mat, right = 0.3 of matmul, label=above:\small$k_n$, label={[xshift=5,yshift=-26]left:\small$k_c$}] {}; 
    	\draw[arrow] (cembed)--(matmul);
    	\draw[arrow] (nembed)--(matmul);
    	\draw[arrow] (matmul)--(m);
    	\node[name=red_max, content, right =0.5 of m, label={[xshift=5,yshift=-26]left:\small$k_c$}] {};
    	\draw[myarrows, draw=orange!90] (m)--(red_max)node[midway, above]{\small $r$};
    	\node[name=max_pool, content, right = 0.5 of red_max, label={[xshift=5,yshift=-26]left:\small $k_c$}] {};
    	\draw[myarrows, draw=gray!80] (red_max)--(max_pool) node[midway, above]{\small $p$};
    	
    	\node[name=prod, op, above right = .25 and 0.3 of max_pool] {$\mathbf{\cdot}$};
    	\draw[arrow] (max_pool)--(prod);
    	\draw[arrow] (cembed)--(prod);
    	\node[name=masked, content, minimum width=10, right=0.5 of prod, label=above:\small $200$, label={[xshift=5,yshift=-26]left:\small $k_c$}] {};
    	\draw[arrow] (prod)--(masked);
    	
    	\node[name=conv_1_1, content, minimum width=10, right= 0.5 of masked, label=above:\small $200$, label={[xshift=5,yshift=-26]left:\small $k_c$}] {};
        \draw[myarrows, draw=blue!50] (masked)--(conv_1_1) node[midway, above]{\small $c$};
    	\node[name=conv_1_2, content, minimum width=10, right= 0.5 of conv_1_1, label=above:\small $200$, label={[xshift=5,yshift=-26]left:\small $k_c$}] {};
    	\draw[myarrows, draw=blue!50] (conv_1_1)--(conv_1_2) node[midway, above]{\small $c$};
    	
    	\node[name=bn_1, content, minimum width=10, right= 0.5 of conv_1_2, label=above:\small $200$, label={[xshift=5,yshift=-26]left:\small $k_c$}] {};
    	\draw[myarrows, draw=green!80!blue] (conv_1_2)--(bn_1)node[midway, above]{\small $b$};
    	
    	\node[name=mxp_1, content, minimum width=10, minimum height=25, label=above:\small $200$, right=0.5 of bn_1, label={[xshift=5,yshift=-16]left:\small $\frac{k_c}{2}$}] {};
    	\draw[myarrows, draw=gray!80] (bn_1)--(mxp_1)node[midway, above]{\small $p$};
    	
    	\node[name=conv_2_1, content, minimum width=10, minimum height=25, label=above:\small $200$, right=0.5 of mxp_1, label={[xshift=5,yshift=-16]left:\small $\frac{k_c}{2}$}] {};
    	\draw[myarrows, draw=blue!50] (mxp_1)--(conv_2_1)node[midway, above]{\small $c$};
    	\node[name=conv_2_2, content, minimum width=10, minimum height=25, label=above:\small $200$, right=0.5 of conv_2_1, label={[xshift=5,yshift=-16]left:\small $\frac{k_c}{2}$}] {};
    	\draw[myarrows, draw=blue!50] (conv_2_1)--(conv_2_2)node[midway, above]{\small $c$};
    	
    	\node[name=bn_2, content, minimum width=10, minimum height=25, right=0.5 of conv_2_2, label=above:\small $200$, label={[xshift=5,yshift=-16]left:\small $\frac{k_c}{2}$}] {};	
    	\draw[myarrows, draw=green!80!blue] (conv_2_2)--(bn_2)node[midway, above]{\small $b$};
    	\node[name=mxp_2, content, minimum width=10, minimum height=12, label=above:\small $200$, right=0.5 of bn_2, label={[xshift=5,yshift=-10]left:\small $\frac{k_c}{4}$}] {};
    	\draw[myarrows, draw=gray!80] (bn_2)--(mxp_2)node[midway, above]{\small $p$};

        \node[name=conv_3_1, content, minimum width=10, minimum height=12, label=above:\small $200$, right=0.5 of mxp_2, label={[xshift=5,yshift=-10]left:\small $\frac{k_c}{4}$}] {};
        \draw[myarrows, draw=blue!50] (mxp_2)--(conv_3_1)node[midway, above]{\small $c$};
        \node[name=conv_3_2, content, minimum width=10, minimum height=12, label=above:\small $200$, right=0.5 of conv_3_1, label={[xshift=5,yshift=-10]left:\small $\frac{k_c}{4}$}] {};
        \draw[myarrows, draw=blue!50] (conv_3_1)--(conv_3_2)node[midway, above]{\small $c$};
        \node[name=bn_3, content, minimum width=10, minimum height=12, label=above:\small $200$, right=0.5 of conv_3_2, label={[xshift=5,yshift=-10]left:\small $\frac{k_c}{4}$}] {};
        \draw[myarrows, draw=green!80!blue] (conv_3_2)--(bn_3)node[midway, above]{\small $b$};
        
        \node[name=res, content, minimum width=10, minimum height=6, label=above:\small $200$, label=right:\small $1$, fill=red!10, right=0.5 of bn_3] {};
        \draw[myarrows, draw=purple!50] (bn_3)--(res)node[midway, above]{\small $\tilde{p}$};
        
        \node[name=legend_0, draw=white, fill=white] at (6,-2) {};
        \node[name=legend_1, draw=white, fill=white] at (6.85,-2) {};
        \draw[myarrows, name=lookup, draw=red!70, label=above:lookup] (legend_0)--(legend_1) node[midway, below]{\small lookup} node[midway, above]{\small $l$};
        
        \node[name=legend_2, draw=white, fill=white] at (7.65,-2) {};
        \node[name=legend_3, draw=white, fill=white] at (8.5,-2) {};
        \draw[myarrows, name=reduce_max, draw=orange!90] (legend_2)--(legend_3) node[midway, below]{\small reduce max}  node[midway, above]{\small $r$};
    	
    	\node[name=legend_4, draw=white, fill=white] at (9.6,-2) {};
    	\node[name=legend_5, draw=white, fill=white] at (10.45,-2) {};
        \draw[myarrows, name=mp, draw=gray!80] (legend_4)--(legend_5) node[midway, below]{\small max pool}  node[midway, above]{\small $p$};

        \node[name=legend_6, draw=white, fill=white] at (11.5,-2) {};
        \node[name=legend_7, draw=white, fill=white] at (12.35,-2) {};
        \draw[myarrows, name=convolution, draw=blue!50] (legend_6)--(legend_7) node[midway, below]{\small convolution}  node[midway, above]{\small $c$};

    	\node[name=legend_8, draw=white, fill=white] at (13.6,-2) {};
    	\node[name=legend_9, draw=white, fill=white] at (14.55,-2) {};
        \draw[myarrows, name=batch_norm, draw=green!80!blue] (legend_8)--(legend_9) node[midway, below]{\small batch norm.}  node[midway, above]{\small $b$};
        
        \node[name=legend_10, draw=white, fill=white] at (15.65,-2) {};
        \node[name=legend_11, draw=white, fill=white] at (16.5,-2) {};
        \draw[myarrows, name=mean_pool, draw=purple!50] (legend_10)--(legend_11) node[midway, below]{\small mean pool}node[midway, above]{\small $\tilde{p}$};
        
        \node[name=place_holder_upper, draw=white, fill=white, below=-.5 of legend_0] {};
        \node[name=place_holder, draw=white, fill=white, below=.4 of legend_0] {};
        
        \node[ dashed, thick, label=below:Legend,minimum height=30, minimum width=320] at (11.4,-2.1) {};

        \node[draw, dashed, thick, minimum width=142, minimum height=130] at (2.90,-0.9) {};
        \node [right] at (0.4,1.65) {\small Content Masking $\tau$};
        
        \node[draw, dashed, thick, minimum width=325, minimum height=70] at (11.2,0.14) {};
        \node [right] at (5.35,1.65) {\small Target Fusion $\xi$};

\end{tikzpicture}    
    \caption{Architecture of the target fusion and relationship-dependent content masking process in ConMask. $k_c$ is the length of the entity description and $k_n$ is the length of the relationship name. This figure is best viewed in color.}
    \label{fig:fcn_mask}
\end{figure*}
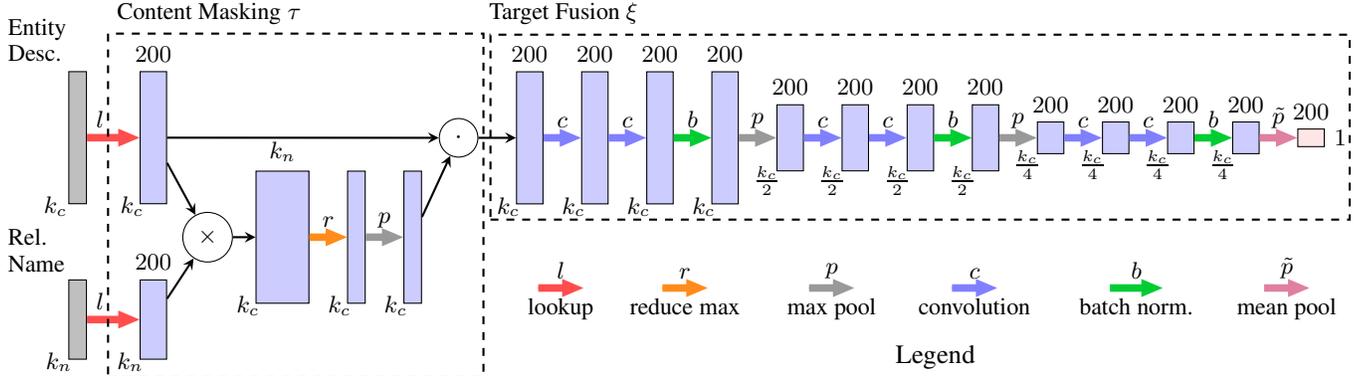

To recap, the relationship-dependent content masking process described here assigns importance weights to words in an entity's description based on the similarity between each word's context and the given relationship. After non-relevant content is masked, the model needs to learn a single embedding vector from the masked content matrix to compare with the embeddings of candidate target entities. 

\subsection{Target Fusion}

Here we describe how ConMask extracts word-based entity embeddings. We call this process the \emph{target fusion} function $\xi$, which distills an embedding using the output of Eq.~\ref{eq:content_masking}. 

Initially, we looked for solutions to this problem in recurrent neural networks (RNNs) of various forms. Despite their popularity in NLP-related tasks, recent research has found that RNNs are not good at performing ``extractive'' tasks~\cite{See2017}. RNNs do not work well in our specific setting because the input of the Target Fusion is a masked content matrix, which means most of the stage inputs would be zero and hence hard to train.

In this work we decide to use fully convolutional neural network (FCN) as the target fusion structure. A CNN-based structure is well known for its ability to capture peak values using convolution and pooling. Therefore FCN is well suited to extract useful information from the weighted content matrix. Our adaptation of FCNs yields the target fusion function $\xi$, which generates a $k$-dimensional embedding using the output of content masking $\tau(\phi(e),$ $\psi(r))$ where $e$ is either a head or tail entity from a partial triple.

Figure~\ref{fig:fcn_mask} shows the overall architecture of the target fusion process and its dependent content masking process. The target fusion process has three FCN layers. In each layer, we first use two $1$-D convolution operators to perform affine transformation, then we apply $sigmoid$ as the activation function to the convoluted output followed by batch normalization~\cite{Ioffe2015} and max-pooling. The last FCN layer uses mean-pooling instead of max-pooling to ensure the output of the target fusion layer always return a single $k$-dimensional embedding. 

Note that the FCN used here is different from the one that typically used in computer vision tasks~\cite{Chen2016}. Rather than reconstructing the input, as is typical in CV, the goal of target fusion is to extract the embedding w.r.t given relationship, therefore we do not have the de-convolution operations. Another difference is that we reduce the number of embeddings by half after each FCN layer but do not increase the number of channels, \ie, the embedding size. This is because the input weighted matrix is a sparse matrix with a large portion of zero values, so we are essentially fusing peak values from the input matrix into a single embedding representing the target entity.

\subsection{Semantic Averaging}

Although it is possible to use target fusion to generate all entity embeddings used in ConMask, such a process would result in a large number of parameters. Furthermore, because the target fusion function is an extraction function it would be odd to apply it to entity names where no extraction is necessary. So, we also employ a simple \emph{semantic averaging} function $\eta(\mathbf{W}) = \frac{1}{k_{l}}\Sigma_{i}^{k_{l}}\mathbf{W}_{[i,:]}$ that combines word embeddings to represent entity names and for generating background representations of other textual features, where $\mathbf{W} \in \mathcal{R}^{k_l\times k}$ is the input embedding matrix from the entity description $\phi(\cdot)$ or the entity or relationship name $\psi(\cdot)$.

To recap: at this point in the model we have generated entity embeddings through the content masking and target fusion operations. The next step is to define a loss function that finds one or more entities in the KG that most closely match the generated embedding.

\begin{table*}[t]
    \caption{Open-world Entity prediction results on DBPedia50k and DBPedia500k. For Mean Rank (MR) lower is better. For HITS@10 and Mean Reciprocal Rank (MRR) higher is better.}
    \label{tab:entity_prediction}
    \centering
    \begin{adjustbox}{max width=\linewidth}
        \begin{tabular}{r ccccccc c ccccccc}
             &  \multicolumn{7}{c@{\quad}}{DBPedia50k} && \multicolumn{7}{c@{\quad}}{DBPedia500k}  \\\cline{2-8}\cline{10-16}
            
             & \multicolumn{3}{c@{\quad}}{Head} && \multicolumn{3}{c@{\quad}}{Tail} && \multicolumn{3}{c@{\quad}}{Head} && \multicolumn{3}{c@{\quad}}{Tail} \\\cline{2-4}\cline{6-8}\cline{10-12}\cline{14-16}
            Model & MR & HITS@10 & MRR && MR & HITS@10 & MRR && MR & HITS@10 & MRR && MR & HITS@10 & MRR\\ \midrule
            
            Target Filtering Baseline & $605$ & $0.07$ & $0.07$ && $104$ & $0.23$ & $0.11$ && $20667$ & $0.01$ & $0.01$ && $3480$ & $0.02$ & $0.01$ \\
            Semantic Averaging & $513$ & $0.11$ & $0.12$ && $93$ & $0.29$ & $0.15$ && $8144$ & $0.04$ & $0.09$ && $1002$ & $0.16$ & $0.13$ \\
            DKRL (2-layer CNN) & $490$ & $0.09$ & $0.08$ && $70$ & $0.40$ & $0.23$ && $19776$ & $0.01$ & $0.01$ && $2275$ & $0.04$ & $0.04$ \\
            \hline
            \textbf{ConMask} & $\mathbf{95}$ & $\mathbf{0.39}$ & $\mathbf{0.35}$ && $\mathbf{16}$ & $\mathbf{0.81}$ & $\mathbf{0.61}$ && $\mathbf{2877}$ & $\mathbf{0.17}$ & $\mathbf{0.33}$ && $\mathbf{165}$ & $\mathbf{0.52}$ & $\mathbf{0.47}$ \\
        \end{tabular}
    \end{adjustbox}
\end{table*}

\subsection{Loss Function} 

To speed up the training and take to advantage of the performance boost associated with a list-wise ranking loss function~\cite{Shi2016proje}, we designed a partial list-wise ranking loss function that has both positive and negative target sampling: 

\begin{equation}
\label{eq:loss}
\mathcal{L}(h,r,t)=\begin{cases}
           \sum\limits_{h_+ \in E^+} -\frac{\log(S(h_+, r, t, E^+\cup E^-))}{|E^+|}, p_c > 0.5\\
           \sum\limits_{t_+ \in E^+} -\frac{\log(S(h, r, t_+, E^+\cup E^-))}{|E^+|}, p_c \leq 0.5\\
                \end{cases},
\end{equation}

\noindent{}where $p_c$ is the corruption probability drawn from an uniform distribution $U[0,1]$ such that when $p_c > 0.5$ we keep the input tail entity $t$ but do positive and negative sampling on the head entity and when $p_c \leq 0.5$ we keep the input head entity $h$ intact and do sampling on the tail entity. $E^+$ and $E^-$ are the sampled positive and negative entity sets drawn from the positive and negative target distribution $P_+$ and $P_-$ respectively. Although a type-constraint or frequency-based distribution may yield better results, here we follow the convention and simply apply a simple uniform distribution for both $P_+$ and $P_-$. When $p_c \leq 0.5$, $P_+$ is a uniform distribution of entities in $\{t_+|\langle h, r, t_+\rangle \in \mathbf{T}\}$ and $P_-$ is an uniform distribution of entities in $\{t_-|\langle h, r, t_-\rangle \notin \mathbf{T}\}$. On the other hand when $p_c > 0.5$, $P_+$ is an uniform distribution of entities in $\{h_+|\langle h_+, r, t\rangle \in \mathbf{T}\}$ and $P_-$ is an uniform distribution of entities in $\{h_-|\langle h_-, r, t\rangle \notin \mathbf{T}\}$. The function $S$ in Eq.~\ref{eq:loss} is the softmax normalized output of ConMask:

\begin{equation}\label{eq:output}
    S(h,r,t,E^\pm) = \begin{cases}
                \frac{\exp(\textrm{ConMask}(h,r,t))} {\sum\limits_{e\in E^\pm}\exp(\textrm{ConMask}(e, r, t))}, p_c > 0.5 \\
                \frac{\exp(\textrm{ConMask}(h,r,t))}{ \sum\limits_{e\in E^\pm}\exp(\textrm{ConMask}(h, r, e))}, p_c \leq 0.5 \\
                \end{cases}.
\end{equation}

Note that Eq.~\ref{eq:loss} is actually a generalized form of the sampling process used by most existing KGC models. When $|E_+|=1$ and $|E_-|=1$, the sampling method described in Eq.~\ref{eq:loss} is the same as the triple corruption used by TransE~\cite{Bordes2013}, TransR~\cite{Lin2015}, TransH~\cite{Wang2014}, and many other closed-world KGC models. When $|E_+| = |\{t|\langle h,r,t\rangle \in \mathbf{T}\}|$, which is the number of all true triples given a partial triple $\langle h$, $r$, ?$\rangle$, Eq.~\ref{eq:loss} is the same as ProjE\_listwise~\cite{Shi2016proje}.

\section{Experiments}\label{sec:kgc_exp}

The previous section described the design decisions and modelling assumptions of ConMask. In this section we present the results of experiments performed on old and new data sets in both open-world and closed-world KGC tasks. 

\subsection{Settings}
Training parameters were set empirically but without fine-tuning. We set the word embedding size $k=200$, maximum entity content and name length $k_c=k_n=512$. The word embeddings are from the publicly available pre-trained $200$-dimensional GloVe embeddings~\cite{Pennington2014}. The content masking window size $k_m=6$, number of FCN layers $k_{fcn}=3$ where each layer has $2$ convolutional layers and a BN layer with a moving average decay of $0.9$ followed by a dropout with a keep probability $p=0.5$. Max-pooling in each FCN layer has a pool size and stride size of $2$. The mini-batch size used by ConMask is $k_b=200$. We use Adam as the optimizer with a learning rate of $10^{-2}$. The target sampling set sizes for $|E_+|$ and $|E_-|$ are $1$ and $4$ respectively. All open-world KGC models were run for at most $200$ epochs. All compared models used their default parameters.

ConMask is implemented in TensorFlow. The source code is available at \url{https://github.com/bxshi/ConMask}.

\subsection{Data Sets}

\begin{table}[ht]
    \caption{Data set statistics. }
    \label{tab:dataset}
    \centering
    \begin{adjustbox}{max width=\linewidth}
    \begin{tabular}{rrrrrr}
                &          &               & \multicolumn{3}{c}{Triples} \\
        \cline{4-6}
        Data set & Entities & Rel. & \multicolumn{1}{c}{Train}& \multicolumn{1}{c}{Validation} & \multicolumn{1}{c}{Test} \\
        \midrule
        FB15k   & $14,951$ & $1,345$       & $483,142$ & $50,000$   & $59,071$ \\
        FB20k   & $19,923$ & $1,345$       & $472,860$ & $48,991$   & $90,149$ \\
        \hline
        DBPedia50k & $49,900$ & $654$      & $32,388$  & $399$      & $10,969$ \\
        DBPedia500k& $517,475$ & $654$     & $3,102,677$ & $10,000$ & $1,155,937$ \\
        \bottomrule
    \end{tabular}
    \end{adjustbox}
\end{table}

\begin{table*}[t]
    \caption{Closed-world KGC on head and tail prediction. For HITS@10 higher is better. For Mean Rank (MR) lower is better.}
    \label{tab:closed_prediction}
    \centering
    \begin{adjustbox}{max width=\linewidth}
        \begin{tabular}{r ccccc c ccccc c ccccc }
             &  \multicolumn{5}{c@{\quad}}{FB15k} && \multicolumn{5}{c@{\quad}}{DBPedia50k} && \multicolumn{5}{c@{\quad}}{DBPedia500k} \\\cline{2-6}\cline{8-12}\cline{14-18}
            
             & \multicolumn{2}{c@{\quad}}{Head} && \multicolumn{2}{c@{\quad}}{Tail} && \multicolumn{2}{c@{\quad}}{Head} && \multicolumn{2}{c@{\quad}}{Tail} && \multicolumn{2}{c@{\quad}}{Head} && \multicolumn{2}{c@{\quad}}{Tail} \\\cline{2-3}\cline{5-6}\cline{8-9}\cline{11-12}\cline{14-15}\cline{17-18}
            Model & MR & HITS@10 &&MR & HITS@10 && MR & HITS@10 && MR & HITS@10 && MR & HITS@10 && MR & HITS@10\\ \midrule
            
            TransE & $189$ & $0.68$  && $92$ & $0.75$ && $2854$ & $0.37$ && $734$ & $0.68$ && $10034$ & $0.15$ && $2472$ & $\mathbf{0.45}$\\
            TransR &  $186$ & $\mathbf{0.71}$ && $87$ & $\mathbf{0.77}$ && $2689$ & $0.39$ && $718$ & $0.67$ && - & - && - & - \\
            \hline
 \textbf{ConMask} & $\mathbf{116}$ & $0.62$ && $\mathbf{80}$ & $0.62$ && $\mathbf{1063}$ & $\mathbf{0.41}$ && $\mathbf{141}$ & $\mathbf{0.72}$ &&  $\mathbf{1512}$ & $\mathbf{0.21}$ && $\mathbf{1568}$ & $0.20$ \\
        \end{tabular}
    \end{adjustbox}
\end{table*}

The Freebase 15K (FB15k) data set is widely used in KGC. But FB15k is fraught with reversed- or synonym-triples~\cite{Toutanova2015} and does not provide sufficient textual information for content-based KGC methods to use. Due to the limited text content and the redundancy found in the FB15K data set, we introduce two new data sets DBPedia50k and DBPedia500k for both open-world and closed-world KGC tasks.\ignore{These data sets are based on DBPedia's April 2016 release combined with the full article text of Wikipedia from October 2016. As in related work, the Wikipedia article text was preprocessed to remove Wiki-markup, tables, and stop words~\cite{Xie2016,Yang2015}. A vocabulary is constructed by collecting words that appear more than five times.} Statistics of all data sets are shown in Tab.~\ref{tab:dataset}. 

The methodology used to evaluate the open-world and closed-world KGC tasks is similar to the related work. Specifically, we randomly selected $90\%$ of the entities in the KG and induced a KG subgraph using the selected entities, and from this reduced KG, we further removed $10\%$ of the relationships, \ie, graph-edges, to create KG$_\textrm{train}$. All other triples not included in KG$_\textrm{train}$ are held out for the test set.\ignore{ Details of the test set are discussed in the task descriptions.}

\begin{table*}[ht]
    \caption{Entity prediction results on DBPedia50k data set. Top-$3$ predicted tails are shown with the correct answer in bold.}
    \label{tab:case_study}
    \centering
    \begin{adjustbox}{max width=\linewidth}
        \begin{tabular}{r r | l}
        Head & Relationship & Predicted Tails \\\midrule
        
        Chakma language & languageFamily &\textbf{Indo-Aryan language}, Hajong language, Language\\ 
        \hline
        
        Gabrielle Stanton & notableWork & Star Trek: Deep Space Nine, A Clear and Present Danger, Pilot (Body of Proof)\\ 
        \hline
        
        MiniD & influencedBy & \textbf{Lua}, Delphi, MorphOS\\ 
        \hline
        
        The Time Machine (1960 film) & writer & Writer, \textbf{David Duncan}, Jeff Martin \\
        \end{tabular}
    \end{adjustbox}
\end{table*}

\subsection{Open-World Entity Prediction}

For the open-world KGC task, we generated a test set from the $10\%$ of entities that were held out of KG$_\textrm{train}$. This held out set has relationships that connect the test entities to the entities in KG$_\textrm{train}$. So, given a held out entity-relationship partial triple (that was not seen during training), our goal is to predict the correct target entity within KG$_\textrm{train}$.

To mitigate the excessive cost involved in computing scores for all entities in the KG, we applied a target filtering method to all KGC models. Namely, for a given partial triple $\langle h$, $r$, ?$\rangle$ or $\langle${?}, $r$, $t \rangle$, if a target entity candidate has not been connected via relationship $r$ before in the training set, then it is skipped, otherwise we use the KGC model to calculate the actual ranking score. Simply put, this removes relationship-entity combinations that have never before been seen and are likely to represent nonsensical statements. The experiment results are shown in Tab.~\ref{tab:entity_prediction}. 

As a naive baseline we include the target filtering baseline method in Tab.~\ref{tab:entity_prediction}, which assigns random scores to all the entities that pass the target filtering. Semantic Averaging is a simplified model which uses contextual features only. DKRL is a two-layer CNN model that generates entity embeddings with entity description~\cite{Xie2016}. We implemented DKRL ourselves and removed the structural-related features so it can work under open-world KGC settings.

We find that the extraction features in ConMask do boost mean rank performance by at least $60\%$ on both data sets compared to the extraction-free Semantic Averaging. Interestingly, the performance boost on the larger DBPedia500k data set is more significant than the smaller DBPedia50k, which indicates that the extraction features are able to find useful textual information from the entity descriptions.

\subsection{Closed-World Entity Prediction}

Because the open-world assumption is less restrictive than the closed-world assumption, it is possible for ConMask to perform closed-world tasks, even though it was not designed to do so. So in Tab.~\ref{tab:closed_prediction} we also compare the ConMask model with other closed-world methods on the standard FB15k data set as well as the two new data sets. Results from TransR are missing from the DBPedia500k data set because the model did not complete training after 5 days.

We find that ConMask sometimes outperforms closed-world methods on the closed-world task. ConMask especially shows improvement on the DBPedia50k data set; this is probably because the random sampling procedure used to create DBPedia50k generates a sparse graph. closed-world KGC models, which rely exclusively on structural features, have a more difficult time with sub-sampled KGs. 

\subsection{Discussion}

In this section we elaborate on some actual prediction results and show examples that highlight the strengths and limitations of the ConMask model. 

Table~\ref{tab:case_study} shows 4 KGC examples. In each case, ConMask was provided the head and the relationship and asked to predict the tail entity. In most cases ConMask successfully ranks the correct entities within the top-$3$ results. \textsf{Gabrielle Stanton}'s \textsf{notableWork} is an exception. Although Stanton did work on \textsf{Star Trek}, DBPedia indicates that her most notable work is actually \textsf{The Vampire Diaries}, which ranked $4^{\textrm{th}}$. The reason for this error is because the indicator word for \textsf{The Vampire Diaries} was ``consulting producer'', which was not highly correlated to the relationship name ``notable work'' from the model's perspective.

Another interesting result was the prediction given from the partial triple $\langle$\textsf{The Time Machine}, \textsf{writer}, ?$\rangle$. The ConMask model ranked the correct screenwriter \textsf{David Duncan} as the $2^{\textrm{nd}}$ candidate, but the name ``David Duncan'' does not actually appear in the film's description. Nevertheless, the ConMask model was able to capture the correct relationship because the words ``The Time Machine'' appeared in the description of \textsf{David Duncan} as one of his major works.

Although ConMask outperforms other KGC models on metrics such as Mean Rank and MRR, it still has some limitations and room for improvement. First, due to the nature of the relationship-dependent content masking, some entities with names that are similar to the given relationships, such as the \textsf{Language}-entity in the results of the \textsf{languageFamily}-relationship and the \textsf{Writer}-entity in the results of the \textsf{writer}-relationship, are ranked with a very high score. In most cases the correct target entity will be ranked above relationship-related entities. Yet, these entities still hurt the overall performance. It may be easy to apply a filter to modify the list of predicted target entities so that entities that are same as the relationship will be rearranged. We leave this task as a matter for future work.

\section{Conclusion and Future Work}\label{sec:kgc_conclusion}

In the present work we introduced a new open-world Knowledge Graph Completion model ConMask that uses relationship-dependent content masking, fully convolutional neural networks, and semantic averaging to extract relationship-dependent embeddings from the textual features of entities and relationships in KGs. Experiments on both open-world and closed-world KGC tasks show that the ConMask model has good performance in both tasks. Because of problems found in the standard KGC data sets, we also released two new DBPedia data sets for KGC research and development.

The ConMask model is an extraction model which currently can only predict relationships if the requisite information is expressed in the entity's description. The goal for future work is to extend ConMask with the ability to find new or implicit relationships.


\small{

\bibliographystyle{aaai}
}

\end{document}